\documentclass[10pt,twocolumn,letterpaper]{article}

\usepackage{cvpr}
\usepackage{times}
\usepackage{epsfig}
\usepackage{graphicx}
\usepackage{amsmath}
\usepackage{amssymb}

\usepackage{booktabs}
\usepackage{array}
\usepackage{setspace}
\usepackage{graphicx}
\usepackage{xcolor}
\usepackage{wrapfig}
\colorlet{myPurple}{blue!40!red}
\usepackage[pagebackref=true,breaklinks=true,letterpaper=true,colorlinks,bookmarks=false]{hyperref}

\makeatletter
\DeclareRobustCommand\onedot{\futurelet\@let@token\@onedot}
\def\@onedot{\ifx\@let@token.\else.\null\fi\xspace}
\def\eg{\emph{e.g}\onedot} 
\def\ie{\emph{i.e}\onedot}

\def\etal{\emph{et al}\onedot}

 \cvprfinalcopy 

\def\cvprPaperID{877} 

\ifcvprfinal\fi
\begin{document}

\title{Error Correction for Dense Semantic Image Labeling}
\author{Yu-Hui Huang$^{1}\footnotemark[1]$ \quad Xu Jia$^{2}\thanks{Equal contribution.} $ \quad Stamatios Georgoulis$^{1}$ \\ Tinne Tuytelaars$^{2}$ \quad Luc Van Gool$^{1,3}$\\
\\
$^{1}$KU-Leuven/ESAT-PSI, Toyota Motor Europe (TRACE) \quad  $^{2}$KU-Leuven/ESAT-PSI, IMEC \\
  $^{3}$ETH/DITET-CVL \\
  }
\maketitle

\begin{abstract}   
Pixelwise semantic image labeling is an important, yet challenging, task with many applications. 
Typical approaches to tackle this problem involve either the training of deep networks on vast amounts of images to directly infer the labels or the use of probabilistic graphical models to jointly model the dependencies of the input (\ie images) and output (\ie labels). 
Yet, the former approaches do not capture the structure of the output labels, which is crucial for the performance of dense labeling, and the latter rely on carefully hand-designed priors that require costly parameter tuning via optimization techniques, which in turn leads to long inference times. 
To alleviate these restrictions, we explore how to arrive at dense semantic pixel labels given both the input image and an initial estimate of the output labels. 
We propose a parallel architecture that: 1) exploits the context information through a LabelPropagation network to propagate correct labels from nearby pixels to improve the object boundaries, 2) uses a LabelReplacement network to directly replace possibly erroneous, initial labels with new ones, and 3) combines the different intermediate results via a Fusion network to obtain the final per-pixel label. 
We experimentally validate our approach on two different datasets for the semantic segmentation and face parsing tasks respectively, where we show improvements over the state-of-the-art.
We also provide both a quantitative and qualitative analysis of the generated results. 
\end{abstract}

\section{Introduction}
\label{sec:intro}



The problem of assigning dense semantic labels to images finds application in many tasks, like indoor navigation, human-computer interaction, image search engines, and VR or AR systems, to name a few. 
The goal is to assign a class label to every pixel, from a pre-defined set of labels. 

In the literature, several methods have been proposed to tackle this problem.
Recently, Deep Convolutional Neural Networks (DCNNs) have become the mainstream for dense semantic image labeling, starting with the Fully Convolutional Network (FCN) proposed by Long \etal \cite{long2015fully, shelhamer2017fully}.
Despite their great representational power, feed-forward DCNN-based approaches tend to produce overly smooth results near the object boundaries 
and do not consider the relations among nearby pixels when predicting the semantic labels.
Different strategies have been proposed to cope with these issues. 
One popular way is to apply probabilistic graphical models, like dense Conditional Random Fields (CRFs) \cite{krahenbuhl2011efficient, chen2016deeplab}, as a post-processing step as is done in \cite{chen2015semantic}.
The pairwise potentials in the CRF impose the consistency of labeling between nearby pixels, and the fully connected CRF delineates the object boundary. 
Although dense CRFs perform well on the refinement of the segmentation results, these pairwise potentials have to be carefully hand-designed in order to model the structure of the output space and it takes quite some parameter hyper-tuning to arrive at a satisfactory result with considerable computation time.

\begin{figure}
\begin{center}
\includegraphics[width=0.47\textwidth]{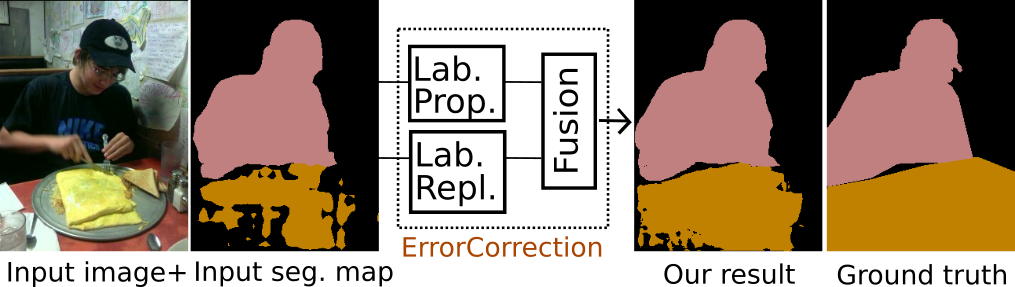}
\end{center}
\caption{The pipeline of the proposed method. Given an input image and a corresponding initial segmentation map, our model predicts a refined segmentation map by implicitly considering the dependencies in the joint space of both the input (\ie images) and output (\ie labels) variables.}
\label{fig:teaser}
\end{figure}

To mitigate these restrictions, we look into ways of achieving the same goal in a more efficient way. 
Our starting point is a variant of the current problem: given an RGB image and an initial estimate of the segmentation map, derived from any dense labeling approach, we seek to estimate a refined segmentation map.
By doing so, we can exploit the dependencies in the joint space of input image and output labels.
We propose a parallel architecture, based on encoder-decoder networks, to deal with the two main sources of error coming from the initialization.
First, a \textit{LabelPropagation} network exploits the context information to predict a pair of displacement vectors $(\Delta x, \Delta y)$ per pixel, \ie a 2D displacement field, in order to propagate labels from nearby pixels to refine the object's shape. 
Obviously, propagating existing labels would not correct cases where the initial labels of all nearby pixels are erroneous and new ones need to be generated. 
In this case, a second \textit{LabelReplacement} network, which runs in parallel with the \textit{LabelPropagation} network, generates new labels directly from the input pair of RGB image and initial segmentation map.
As a final stage, a \textit{Fusion} network combines the results of these parallel branches 
by predicting a mask to obtain the optimal label for each pixel.
Fig.~\ref{fig:teaser} gives an overview of our pipeline.

\begin{figure*}
\begin{center}
\includegraphics[width=1.00\textwidth]{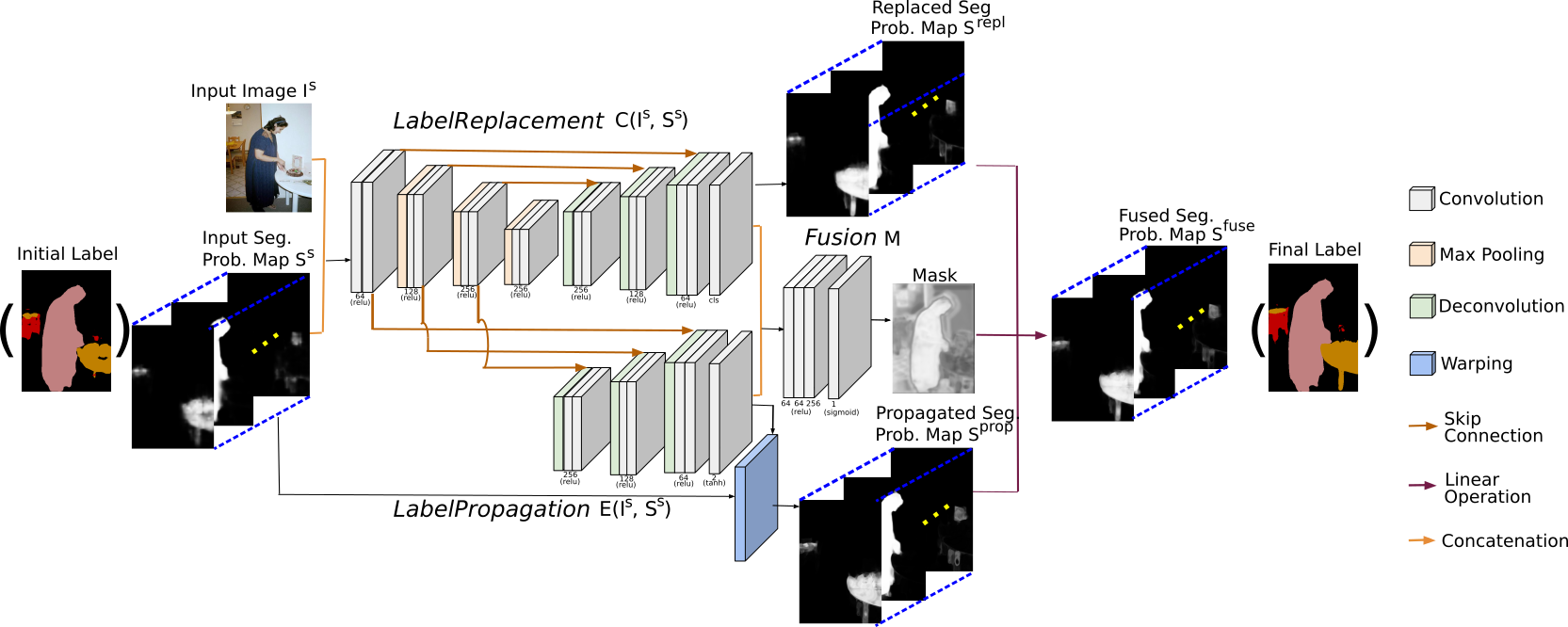}
\end{center}
\caption{The architecture of our pipeline. The \textit{LabelPropagation} network $E$ propagates probability distributions from nearby pixels to refine the object boundaries. In parallel, the \textit{LabelReplacement} network $C$ predicts a new segmentation probability map directly from the input pair of RGB image and initial segmentation map. Finally, the \textit{Fusion} network $M$ combines the results of these branches with a predicted mask to obtain the optimal labeling. The image in the parenthesis denotes the colored label map.}
\label{fig:pipe}
\end{figure*}



Our contributions can be summarized as follows:  
(1) We introduce an efficient post-processing technique for error correction in dense labeling tasks, that works on top of any existing dense labeling approach. 
(2) We propose an end-to-end pipeline that employs different correction strategies by propagating correct labels to nearby pixels (\textit{LabelPropagation} network), replacing the erroneous labels with new ones (\textit{LabelReplacement} network) and fusing the intermediate results (\textit{Fusion} network) in a multi-task learning manner. 
Different from other work \cite{gidaris2016detect}, our method tackles the problem in a parallel rather than sequential way. 
(3) We show that our model is able to improve two state-of-the-art models for the object semantic segmentation task of PASCAL VOC 2012. 
Moreover, our method also improves the performance on a face parsing task.

The paper is organized as follows. 
Sec.~\ref{sec:relatedWork} positions our work w.r.t. earlier work.
Sec.~\ref{sec:method} describes the proposed architecture for error correction in dense semantic labeling.
Experimental results are presented in Sec.~\ref{sec:exp}.
Sec.~\ref{sec:conclusion} concludes the paper.

\section{Related Work}
\label{sec:relatedWork}
The literature on dense semantic labeling is substantial. 
We consider three main categories of related papers.   

\textbf{Deep learning}
The great success of deep learning techniques, such as DCNNs \cite{lecun1998gradient}, in the image classification and object recognition tasks \cite{krizhevsky2012imagenet, simonyan2014very, szegedy2015going} has motivated researchers to apply the same techniques for dense labeling tasks, like semantic segmentation.
First, Long \etal \cite{long2015fully, shelhamer2017fully} transformed existing classification CNN models into FCNs by replacing fully connected layers with convolutional ones such that the network can output label maps.
Next, Badrinarayanan \etal \cite{badrinarayanan2015segnet} proposed an encoder-decoder architecture with skip connections to up-sample the low-resolution feature maps to pixel-wise predictions for segmentation. 
Many following works explore to include more context knowledge. On the one hand local information is important for pixel-level accuracy; on the other hand integrating information from global context can help with local ambiguities.
The most characteristic works involve the use of dilated convolutions \cite{yu2015multi, chen2016deeplab, paszke2016enet}, multi-scale prediction \cite{eigen2015predicting, roy2016multi, zhao2016pyramid}, attention models \cite{Chen2016attention, harley2017segmentation} and feature fusion \cite{liu2015parsenet, pinheiro2016learning}.
Despite great representation power of DCNNs, 
their inability to capture the structure of the output labels affects the performance of dense labeling tasks, especially near the object boundaries.
In particular, feed-forward DCNNs do not explicitly consider the relations among nearby pixels in a local neighborhood of the label space. 

\textbf{Probabilistic graphical models}
As explained above, the DCNNs' inherent invariance to spatial transformations also limits their spatial accuracy in semantic segmentation tasks.
A second line of work explicitly handles this inability by trying to jointly model the dependencies of both the input (\ie images) and the output (\ie labels) variables. 
The most common approach is to apply CRFs \cite{lafferty2001conditional} as a post-processing stage of a DCNN.
The DeepLab models 
 \cite{chen2015semantic, chen2016deeplab} use the fully connected pairwise CRF by Kr{\"a}henb{\"u}hl and Koltun \cite{krahenbuhl2011efficient} to refine the segmentation result of the DCNN. 
Their models incorporate prior knowledge about the structure of the output space in the pairwise potential term to enforce consistency among neighboring or "similarly-looking" pixels. 
In general, in all CRF-based approaches the pairwise potentials have to be carefully hand-designed in order to model the structure in the output space and it takes expensive parameter hyper-tuning to arrive at a satisfactory inference.
Another relevant work is CRF-RNN~\cite{zheng2015conditional}, which uses a neural network to approximate the dense CRF inference process to obtain a good semantic segmentation result. 
However, their model still requires considerable time to do inference.

\textbf{Error correction}
Most recently, a third line of work goes beyond the restrictions imposed by DCNNs and CRFs and tries to model the joint space of input and output variables.
These approaches solve a variant of the traditional dense labeling task: given the input image and an initial estimate of the output labels a network is trained to predict new refined labels, thus being implicitly enforced to reason about the joint input-output space.
These methods come in two flavors, the transform-based ones \cite{yu2015multi, li2016iterative} that learn to directly predict new labels from the initial estimate, and the residual-based ones \cite{carreira2016human} that estimate residual corrections which are added to the initial estimate. 
Gidaris and Komodakis \cite{gidaris2016detect} combined these two flavors for the dense disparity estimation task by proposing a sequential DCNN architecture that is end-to-end trainable. 
Their approach detects the errors in the initial labels, then replaces the incorrect labels with new ones, and finally refines the labels by predicting residual corrections.
Although this method provides good results for improving the continuous values in the dense disparity estimation task, its residual correction stage is difficult to apply to discrete, dense labeling tasks such as semantic segmentation.
Different from that method, we elaborate two branches that account for different types of errors: one for propagating existing labels from nearby pixels and the other for predicting new labels. Finally a fusion module is added to take advantage of both branches.
Moreover, these two branches run in parallel instead of sequentially, thus allowing for faster inference times.

\section{Our Approach}
\label{sec:method}
Given an input RGB image $I^s$ and an initial segmentation probability map $S^s$, we propose an end-to-end pipeline for error correction (see Fig.~\ref{fig:pipe}) which is built upon three networks, \ie the \textit{LabelPropagation}, \textit{LabelReplacement} and \textit{Fusion} networks.
This section provides the details.

\subsection{LabelPropagation network}

We propose to estimate a displacement vector $(\Delta x, \Delta y)$ for each pixel, \ie a 2D displacement field, in order to propagate labels from nearby pixels. 
A warping layer is followed to apply the estimated displacements in order to arrive at an improved segmentation probability map. 
Inspired by \cite{zhou2016-vsaf}, we adopt an encoder-decoder architecture with skip connections for the displacements estimation, which is denoted as \textit{LabelPropagation} network $E$. 
Our work resembles flow-based networks \cite{zhou2016-vsaf, liu2017video}, but unlike those our network learns to predict the displacements from the joint space of both the input and the output variables instead of finding correspondences among different views.

To sum up, given an input image $I^s$ and the initial segmentation probability map $S^s$ our goal is to train a network $E$ that computes an improved segmentation probability map $S^{prop}$ 
by re-sampling $S^s$ according to the predicted 2D displacement field. 
It can be formulated as minimizing the loss function between $S^{prop}$ and the ground truth segmentation map $S^{gt}$,
\begin{equation}
	\begin{aligned}
		\mathcal{L}_{prop} = &\frac{1}{|\mathcal{D}|} \sum_{<I^s,S^s,S^{gt}> \in \mathcal{D}} \mathcal{L}(S^{gt}, E(I^s,S^s)), \\
	\end{aligned}
\end{equation}
where $\mathcal{D}$ is the training dataset, $E(\cdot)$ refers to the \textit{LabelPropagation} network whose parameters we aim to optimize, and $\mathcal{L}$ denotes the cross-entropy loss. 

The \textit{LabelPropagation} network $E$ aims at leveraging the context information from the probability distribution of nearby pixels to predict a pair of displacement vectors $(\Delta x, \Delta y)$, one for each direction, such that a pixel's probability distribution can be re-estimated with respect to its neighbors. 
Here, $(\Delta x, \Delta y)$ denotes the displacement vectors where the model samples the probability distribution from. 
For every pixel $(x_i, y_i)$ in $S^s$, the coordinates w.r.t. the ones after propagation $(x_i^{prop}, y_i^{prop})$ are associated as, 
\begin{equation}
x_i^s = x_i^{prop} - \Delta x_i, y_i^s = y_i^{prop} - \Delta y_i.
\end{equation}

\begin{figure}
\begin{center}
\includegraphics[width=0.48\textwidth]{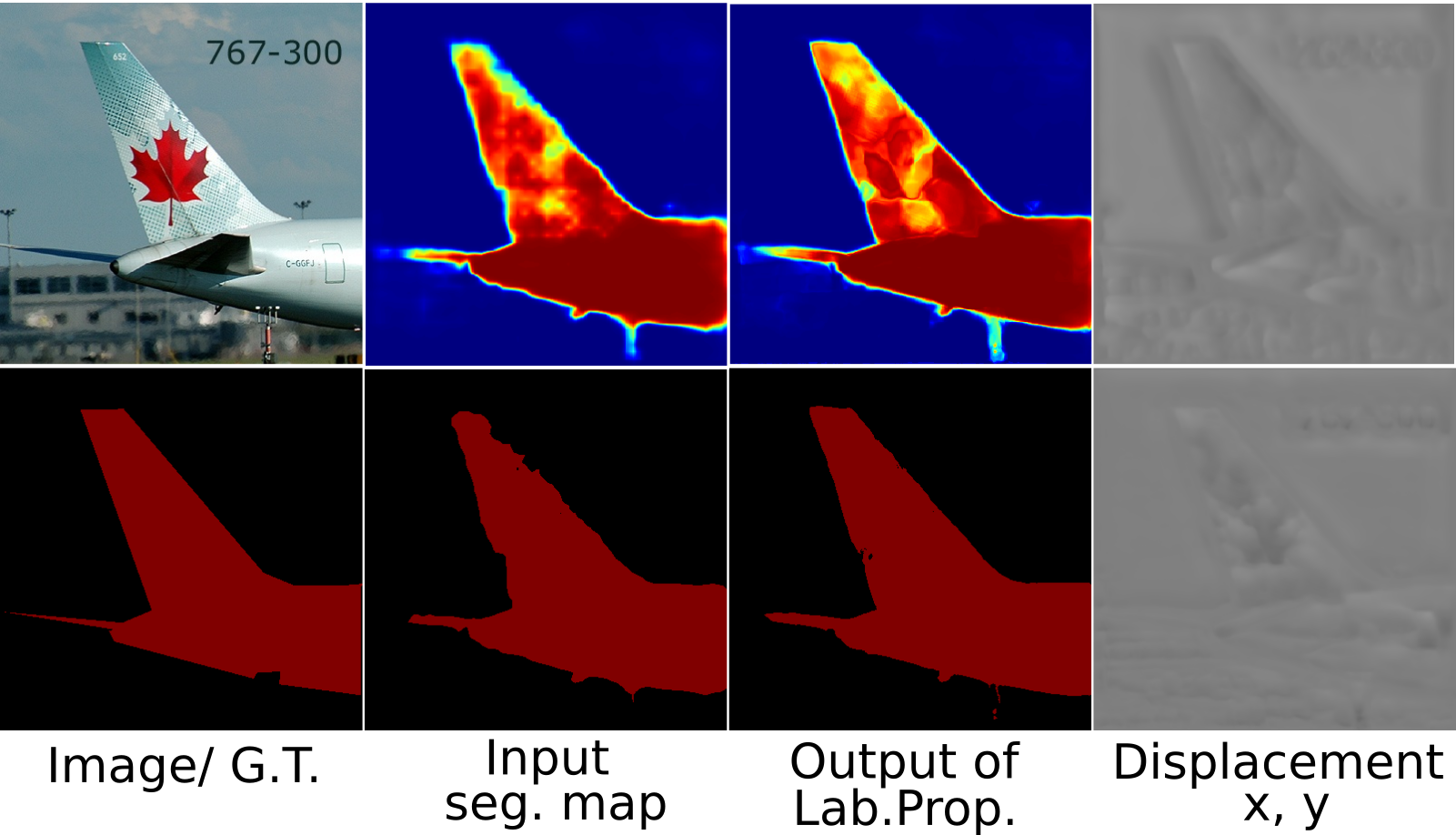}
\end{center}
   \caption{Example result of the \textit{LabelPropagation} network. The first row of the second and third column illustrate the probability map after softmax. The forth column visualizes the predicted displacement vector.}
\label{fig:prop}
\end{figure}

Finally, the initial probability map $S^s$ is warped according to the estimated displacement vectors to generate the refined probability map $S^{prop}$. 
Regarding the warping operation, we use the bilinear sampling kernel in the same way as in \cite{jaderberg15spatial} to allow for end-to-end training,
\begin{equation}
	S_i^{prop} = \sum_{k \in N(x_i^s, y_i^s)} S^s_k(1-|x_i^s - x_k^s|)(1-|y_i^s - y_k^s|),
\end{equation}
where $S_i^{prop}$ denotes the value of the $i$-th pixel at $(x_i^{prop}, y_i^{prop})$ in the output $S^{prop}$, and $N(x_i^s, y_i^s)$ is the 4-neighborhood region of the pixel at $(x_i^s, y_i^s)$ in the input $S^s$.
Its gradients w.r.t. the parameters for displacement estimation can be efficiently computed as in \cite{jaderberg15spatial}. An example of the 2D displacement fields and the output of the \textit{LabelPropagation} network can be seen in Fig.~\ref{fig:prop}.

\subsection{LabelReplacement network}

As explained in the previous section, the \textit{LabelPropagation} network $E$ is able to correct the segmentation error by propagating the possibly correct labels into their neighborhood. 
However, it fails to correct the labels when almost all pixels in a region have initially wrong labels.
To deal with this case, we propose to feed both the input image $I^s$ and the initial segmentation probability map $S^s$ into a fully convolutional \textit{LabelReplacement} network $C$ to directly re-compute a new segmentation probability map $S^{repl}$. 
The network re-estimates a probability vector for each pixel, but this time based on both its appearance and the probability distribution 
of its neighbors. 
Following the same encoder-decoder architecture as in our \textit{LabelPropagation} network, we replace the last layer of the \textit{LabelPropagation} network with a convolutional layer to output the new segmentation probability map. 

In short, given an image $I^s$ and its corresponding initial segmentation probability map $S^s$, we train a network \textit{LabelReplacement} network $C$ to predict a new segmentation probability map $S^{repl}$ based on the initial one $S^s$.
The task can be formulated as minimizing the cross-entropy loss between the newly generated segmentation map $S^{repl}$ and corresponding ground truth labels $S^{gt}$,
\begin{equation}
\mathcal{L}_{repl}= \frac{1}{|\mathcal{D}|}\sum_{<I^s,S^s,S^{gt}> \in \mathcal{D}} \mathcal{L}(S^{gt},C(I^s,S^s)).
\end{equation} 

\subsection{Fusion network}

The \textit{LabelPropagation} and \textit{LabelReplacement} networks work in parallel and are specialized at correcting different types of errors. 
On the one hand, the \textit{LabelPropagation} network $E$ takes into account the nearby pixels and their corresponding class probabilities to propagate the probability vector based on the appearance similarity. 
On the other hand, the \textit{LabelReplacement} network $C$ re-estimates the class labels pixel by pixel. 
To get the best of both worlds, we combine the outputs of these two parallel branches using a \textit{Fusion} network $M$, and train the whole architecture jointly. 
Since the two branches complement each other, our combined model can benefit from a joint training by enforcing each branch to focus on what they are specialized at and leave for the other branch what they are not good at. 
The overall pipeline, including all three networks, can be found in Fig.~\ref{fig:pipe}. 

Design-wise, we use a shared encoder to learn features for both sub-tasks, \ie the \textit{LabelPropagation} and \textit{LabelReplacement} networks, and to also reduce the total number of parameters to be optimized. 
The network then splits into two different decoders in a branched manner, one for predicting the displacement and the other for directly predicting new labels. 
At the final stage, to combine the intermediate results from the two branches, we add the \textit{Fusion} network $M$ that takes 
those intermediate results as input, and predicts a mask $m$ to generate the final segmentation result. 
%
The final result is then computed as a weighted average of the two branches' output in pixel-level,
\begin{equation}
	S^{fuse}= m \odot S^{prop}+(1-m) \odot S^{repl},
\end{equation}
where $S^{prop}$ and $S^{repl}$ are the intermediate segmentation probability maps of the two branches and $\odot$ denotes element-wise multiplication.
Now, the overall loss function can be formulated as:
\begin{equation}
\mathcal{L}_{fuse}= \mathcal{L}(S^{gt}, S^{fuse}) +\mathcal{L}(S^{gt}, S^{prop})+\mathcal{L}(S^{gt}, S^{repl}).
\end{equation}


\subsection{Network architecture}

The \textit{LabelPropagation} and \textit{LabelReplacement} networks share the base architecture, which is based on fully-convolutional encoder-decoders.  
For the encoder, there are four blocks with each one containing two convolutional layers with kernel size 3x3 and a max pooling layer. 
For the decoders, there are three blocks containing one bilinear upsampling layer and two convolutional layers with kernel size 3x3. 
We add three skip connections at the beginning of the three blocks to incorporate information from different resolutions. 
This has been shown to be helpful in producing more details in the decoding process \cite{mao2016-superres, liu2017video}. 
The \textit{Fusion} network predicts a mask to combine both the \textit{LabelPropagation} and \textit{LabelReplacement} networks.
It has three convolutional layers with kernel size 3x3 and another convolutional layer to generate the one-channel mask.
More details on the network hyper-parameters (\eg feature map size, number of channels) can be found in the supplementary material.


\subsection{Training details}

Regarding the training details, we initialize the weights in our networks with Xavier initialization. 
To learn the network parameters, we adopt the ADAM optimizer \cite{Kingma14adam} with a learning rate of 0.0001, $\beta_1$ = 0.9, $\beta_2$ = 0.999 and a batch size of 8. 
The overall training procedure includes about 20,000 iterations. 
For data augmentation, we adopt random mirror, resize between 0.5 and 1.5 for all datasets, and crop to a fixed size according to each dataset. 
The input image is then normalized to [-1,1] and the corresponding initial segmentation probability map is applied using the softmax operation.

\section{Experiments}
\label{sec:exp}
To demonstrate the effectiveness of the proposed method, we evaluate it on two dense labeling tasks, that is, object semantic segmentation and face parsing. 
We also analyze the influence of each component
by comparing their performance when trained independently.
In the following experiments we apply our models on top of semantic segmentation results of several state-of-the-art approaches. 

\subsection{Datasets}

To evaluate the performance of our method on the object semantic segmentation task, we choose the PASCAL VOC2012 segmentation benchmark \cite{everingham2010pascal} as our testbed. 
In total, it consists of 20 classes plus the background class. 
The performance is measured in terms of mean intersection-over-union (IoU). 
Including the additional augmented set annotated by \cite{Hariharan11semantic}, there are 10,582 images for training, 1,449 images for validation and 1,456 images for testing. 

Regarding the face parsing task, we work on the HELEN dataset \cite{Smith2013exemplar}. 
It contains 11 classes, representing different facial components: eyes, eyebrows, nose, and so on. 
In total, there are 2,300 images, divided into a training set with 2,000 images, a val set with 300 images and a test set with 100 images. 
To measure the performance of the different approaches on this task, we adopt the F-measure metric following the protocols previously defined in \cite{liu2015multi, Smith2013exemplar}.

\subsection{PASCAL VOC 2012}

For the PASCAL VOC2012 segmentation benchmark, we apply our networks on top of two state-of-the-art models, DeepLab v2-ResNet (multi-scale) \cite{chen2016deeplab} and PSPNet (single-scale) \cite{zhao2016pyramid}. 
In particular, we first run the inference of these two models to obtain the initial segmentation probability map on the train+aug and val sets.
Note that, these two models were trained on the training set without finetuning on the val set\footnote{The models are provided by the authors.}. 
Using the image and corresponding initial segmentation probability map as input, we train our models on the training set and evaluate them on the val set. 

\begin{table}
\caption{Results of applying our error correction models on top of DeepLabv2-ResNet on the PASCAL VOC 2012 val set.}
\begin{center}
\begin{tabular}{|l |l |l|}
\hline
Method & Training & mIoU \\
\hline\hline
 Deeplab v2-ResNet (multi-scale) & independently & 76.5 \\
 + Dense CRF \cite{chen2016deeplab} & & 77.7 \\
\hline
 +LabelPropagation (ours) & independently & 77.9  \\
 +LabelReplacement (ours) & independently & 77.0 \\
 +Full model (ours) & jointly & \textbf{78.2} \\
\hline
\end{tabular}
\end{center}
\label{tab:deeplab_val}
\end{table}

Table~\ref{tab:deeplab_val} summarizes the results of our ablation study. 
Here, the different proposed networks are trained independently and applied on top of the DeepLab v2-ResNet segmentation result. 
From this table, we can see that adding only the \textit{LabelPropagation} network on top of DeepLab brings 1.4\% improvement compared to the baseline, while adding only the \textit{LabelReplacement} network brings 0.5\% improvement. 
When we train the \textit{LabelPropagation} and \textit{LabelReplacement} networks together with the \textit{Fusion} network, which from now on denotes our full model, this brings the highest improvement, 1.7\%.

\begin{figure}
\begin{center}
\includegraphics[width=0.49\textwidth]{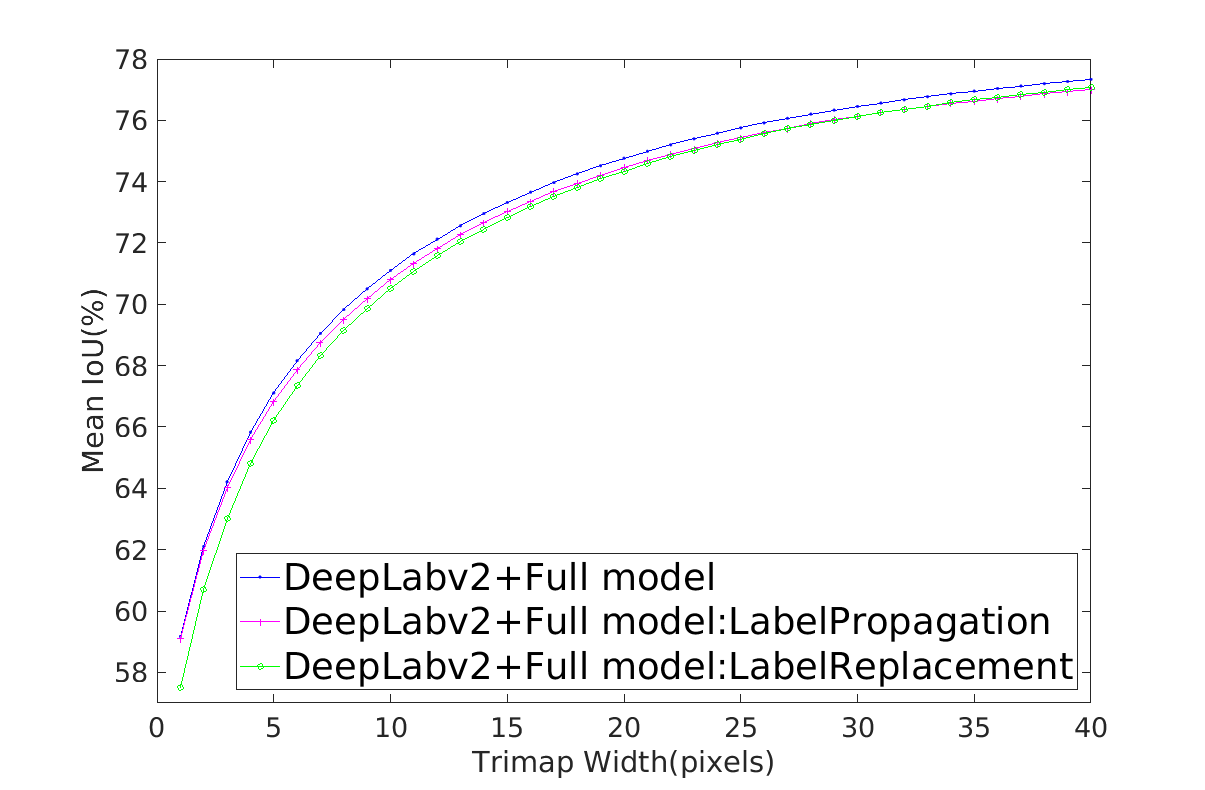}
\end{center}
   \caption{Trimap plot of our full model and its intermediate branches, 
    \ie \textit{LabelPropagation} and \textit{LabelReplacement} networks, on PASCAL VOC 2012.}
\label{fig:trimap_fuse}
\end{figure}

So far, we have evaluated the performance of the \textit{LabelPropagation} and \textit{LabelReplacement} networks when trained independently.
Next, we investigate
the intermediate results generated by these two networks when training them jointly with the \textit{Fusion} network.
In this case, the \textit{LabelPropagation} network scores 77.8\% while the \textit{LabelReplacement} network scores 78.0\%.
For joint training, the \textit{LabelReplacement} network shows 1\% improvement compared to an independent training, but the performance of the \textit{LabelPropagation} network remains roughly the same.
The improvement of our full model
is 1.7\%
compared to the baseline.
We conclude that a joint training brings further improvement compared to an independent training.

Since the \textit{LabelPropagation} and \textit{LabelReplacement} networks complement each other, we hypothesize that we benefit from their joint training by enforcing each network to focus on what they are specialized at and leave for the other network what they are not good at. 
To prove this point, 
we show the trimap result
in Fig.~\ref{fig:trimap_fuse} which quantifies the performance at the object boundary region (details of trimap are described in Sec.~\ref{sec:err_analysis}). 
It shows that the \textit{LabelPropagation} branch outperforms the \textit{LabelReplacement} branch at pixels near the boundary region, which indicates that our full model indeed relies more on this branch for the object's boundaries.
When we train the two networks jointly, the \textit{LabelPropagation} branch focuses on the object boundaries, and as such the \textit{LabelReplacement} branch can pay less attention to these regions where it does not perform well and put more emphasis on the object's inner part.
 

Visualizations of the segmentation results using our full model, and baseline models (DeepLabv2-ResNet and DeepLabv2-ResNet + Dense CRF), can be found in Fig.~\ref{fig:vis_b_mask}.
In the first row, the \textit{table} and \textit{chair} at the right bottom corner are better delineated after applying our method. 
Also, the \textit{tv screen} at the second row becomes more rectangular compared to the baseline methods.
In the third and fourth row, the \textit{leg} of the table and the human respectively are better formed, coming closer to ground truth.
%
\begin{table}
\caption{Results of applying our error correction models on top of PSPNet on the PASCAL VOC 2012 val set.}
\begin{center}
\begin{tabular}{|l|l|l|}
\hline
Method & Training & mIoU \\
\hline
PSPNet (single-scale) \cite{zhao2016pyramid}& independently & 80.7 \\
\hline
+LablePropagation (ours) & independently & 80.3  \\
+LabelReplacement (ours) & independently & 80.8 \\
+Full model (ours)& jointly & \textbf{81.0} \\
\hline
\end{tabular}
\end{center}
\label{tab:psp_psp}
\end{table}
%
Table~\ref{tab:psp_psp} presents the segmentation results on top of PSPNet's result.
The PSPNet's results are already quite good, and as such error correction models like ours tend to bring only marginal improvements. 

Regarding the performance on the test set of PASCAL VOC 2012, we directly apply our networks (full model) on top of precomputed semantic segmentation results on the test set. 
Table~\ref{tab:voc_test} summarizes the per-class performance of the compared methods based on the DeepLab-v2-ResNet and PSPNet models. 
For DeepLab-v2-ResNet, adding Dense CRF brings the performance from 79.1\% up to 79.7\%, while adding our full model further improves it to 80.4\%. 
For PSPNet, our full model performs slightly better than the baseline. 
Nevertheless, our method scores better or equal to PSPNet at 16 out of the 20 classes.

\begin{table*}
\caption{Quantitative results in per-class IoU on the PASCAL VOC 2012 test set.}
\begin{center}
\scalebox{0.68}{
\begin{tabular}{|l|c c c c c c c c c c c c c c c c c c c c c |}
\hline
Method & aero & bike & bird & boat & bottle & bus & car & cat & chair & cow & table & dog & horse & mbike & person & plant & sheep & sofa & train & tv & mIoU \\
\hline
Deeplab & 91.5 & 58.8 & 90.3 & 64.6 & 76.0 & 94.3 & 88.6 & 91.4 & 33.6 & 87.5 & 67.3 & 88.3 & 91.3 & 86.6 & 86.2 & 61.7 & 87.9 & 58.8 & 86.0 & 73.5 & 79.1\\
+dense CRF \cite{chen2016deeplab} & 92.6 & 60.4 & 91.6 & 63.4 & 76.3 & 95.0 & 88.4 & 92.6 & 32.7 & 88.5 & 67.6 & 89.6 & 92.1 & 87.0 & 87.4 & 63.3 & 88.3 & 60.0 & 86.8 & 74.5 & 79.7\\
Ours (+full model) & 92.9 & 63.2 & 91.8 & 66.7& 77.3 & 95.4& 89.1 & 92.3 & 35.4 & 88.0 & 69.5 & 89.1 & 92.3 & 87.2 & 87.3 & 63.3 & 88.6 & 61.8 & 86.6 &75.1 & 80.4\\
\hline\hline
PSPNet \cite{zhao2016pyramid}& 95.8 & 72.7 & 95.0 & 78.9 & 84.4 & \textbf{94.7} & 92.0 & 95.7 & 43.1 & 91.0& 80.3 & 91.3 & 96.3 & \textbf{92.3} & 90.1 & \textbf{71.5} & 94.4 & \textbf{66.9} & 88.8 & 82.0 & 85.4 \\
Ours (+full model) & 95.8 & \textbf{73.1} & \textbf{95.5} & \textbf{79.1} & \textbf{84.5} & 93.1 & \textbf{92.8} & \textbf{96.0} & \textbf{43.3} & \textbf{92.4} & 80.3 & \textbf{91.8} & \textbf{96.4} & 92.1 & \textbf{90.5} & 70.6 & 94.4 & 65.7 & \textbf{89.1} & \textbf{82.6} & \textbf{85.5}\\
\hline
\end{tabular}
}
\end{center}
\label{tab:voc_test}
\end{table*}

\begin{table*}
\caption{Quantitative results on the HELEN dataset using the F-measure metric following the label definition in \cite{liu2015multi}.}
\begin{center}
\begin{tabular}{|l|c c c c c c c c c|}
\hline
Method & eyes & brows & nose & in mouth & upper lip & lower lip & mouth all & facial skin & overall \\
\hline
Liu \textit{et al.} \cite{facemodel} & 72.82 & 69.17 & 91.25 & 77.54 & 65.02 & 74.46 & 86.75 & \textbf{92.07} & 83.96\\
\hline
Ours (+full model)& \textbf{73.97} & \textbf{70.26} & \textbf{93.02} & \textbf{78.65} & \textbf{70.79} & \textbf{77.30} & \textbf{89.01} & 91.97 & \textbf{85.90} \\
\hline
\end{tabular}
\end{center}

\label{tab:helen}
\end{table*}

Compared to DeepLab-v2-CRF, our full model scores 0.5\% and 0.7\% higher on the PASCAL VOC val and test set, respectively.
In terms of speed, the 10-iteration mean-field dense CRF implementation takes 2,386 ms/image on average on an Intel i7-4770K CPU, while our full model takes 396 ms/image on average on an NVIDIA Titan-X GPU, which is about six times faster than dense CRF. 
In addition to efficiency, our model is possible to be plugged into any deep segmentation network for end-to-end training. 
There is no extra hyper-parameter tuning specific to our model.
%

\begin{figure}
\begin{center}

\includegraphics[width=0.45\textwidth]{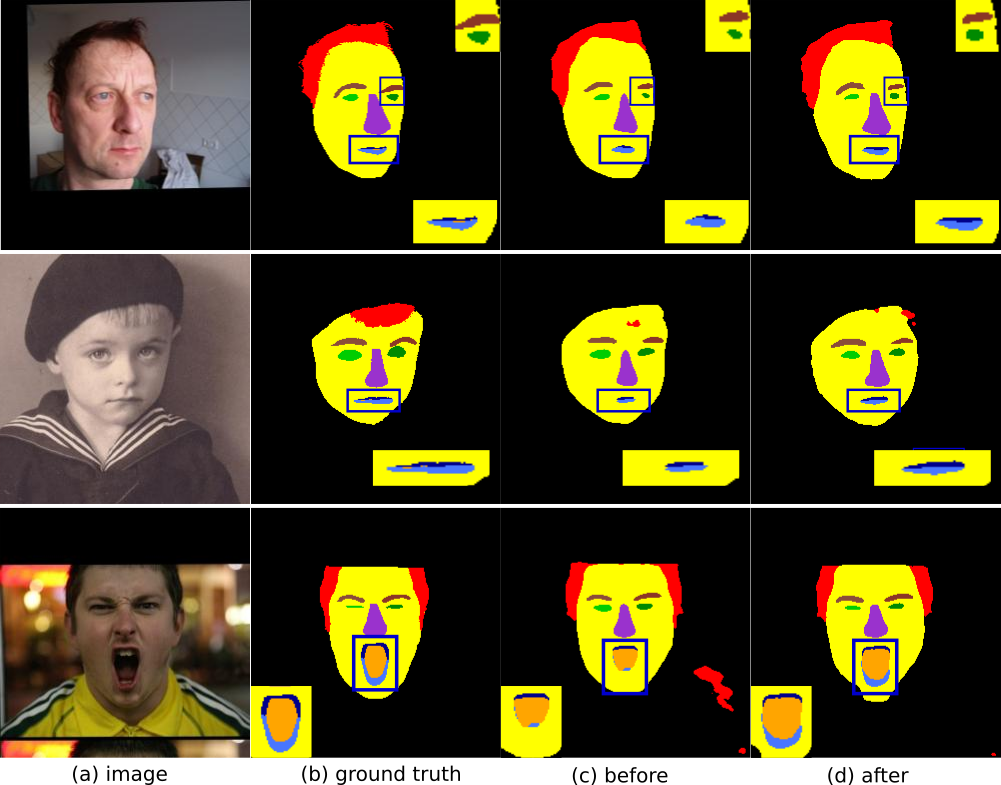}
\end{center}
   \caption{Face parsing results on Helen dataset. For each row, we present (a) input image , (b) ground truth, (c) baseline result, (d) result after applying our model.
    }
\label{fig:helen}
\end{figure}

\subsection{HELEN Dataset}

For these experiments, we follow \cite{liu2015parsenet} and use the resized version of the images from the HELEN dataset. 
The image height and width is about 300 and 500 pixels, respectively. 
As a pre-processing step, we align the images to a canonical position based on the detected five-point facial landmarks.
To derive the five facial keypoints, we first apply the face detector from \cite{Sun2013deep} and obtain the bounding box for the face region. 
For images containing multiple faces, we select the bounding box with the highest overlap with the facial region from the ground truth. 
For images where detection failed, we manually draw the bounding box around the face. 
After getting the face bounding boxes, we apply the 68-point facial landmark detection from \cite{dlib09} and transform those 68 points to 5 landmark points using the script provided by \cite{facemodel}.

As a baseline, we run the inference of the pre-trained model for face parsing from \cite{facemodel} on the training set of the HELEN dataset with an input size of 128 by 128, and upsample the resulted segmentation map to 250 by 250 using bilinear interpolation. 
The baseline network is composed of five consecutive groups of convolutional, ReLU, max pooling (with stride 2) layers in the encoder and a decoder with symmetric configuration but with max pooling layers replaced by bilinear upsampling layers and added skip connections from the encoder to the decoder.

We follow the evaluation criteria used in \cite{liu2015multi, Smith2013exemplar} and report the performance as F-measure for the grouped facial components \textit{eyes}, \textit{brows}, and \textit{in mouth} following the label definition in \cite{liu2015multi, Smith2013exemplar}. 
The overall F-measure of the baseline method is 83.96\%. 
We apply our full model on top of this result to further improve the face parsing performance.

Table~\ref{tab:helen} presents the quantitative results on the HELEN dataset. 
On almost every category, except \textit{facial skin}, our full model brings further improvements compared to the baseline. 
The overall F-measure in our case is 85.90\%, which is about 2\% better compared to the baseline. 
From Fig.~\ref{fig:helen}, we find that the result of the baseline is already quite good. 
Yet, after applying our method the shape of some parts, like the eyes and the mouth, look much closer to the ground truth. 
Moreover, the lower lip in the third row is more complete with our model.

\begin{figure}
\begin{center}
\includegraphics[width=0.48\textwidth]{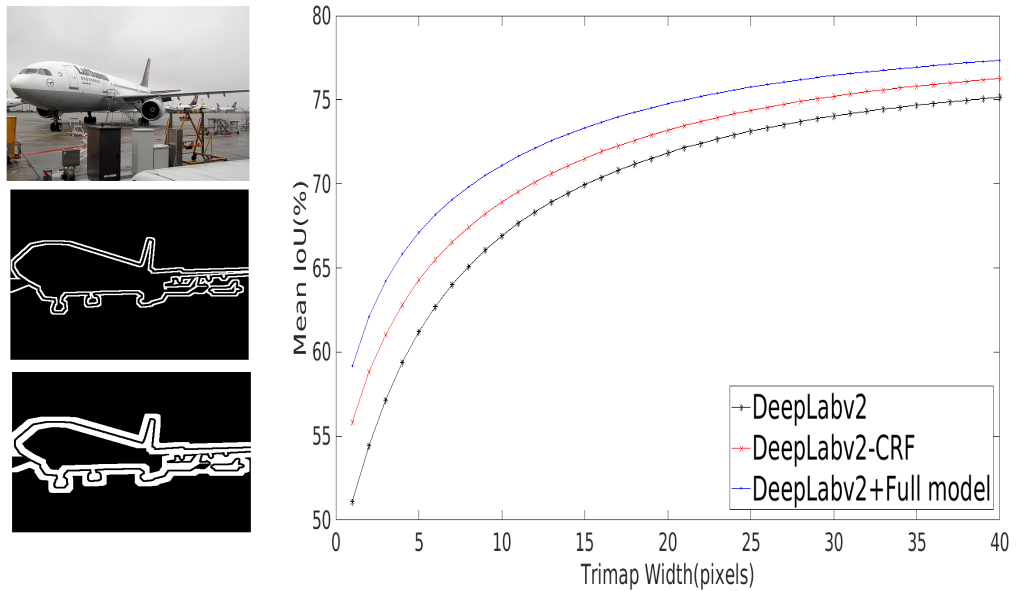}
\end{center}
   \caption{
Performance in mean IoU near object boundaries('trimap'). The left side illustrates two examples of trimap size in three (middle) and in ten (bottom). The right side shows the mean IoU at different trimap sizes from 1 to 40 on PASCAL VOC 2012.}
\label{fig:trimap}
\end{figure}

\begin{figure*}
\begin{center}
\includegraphics[width=0.99\textwidth]{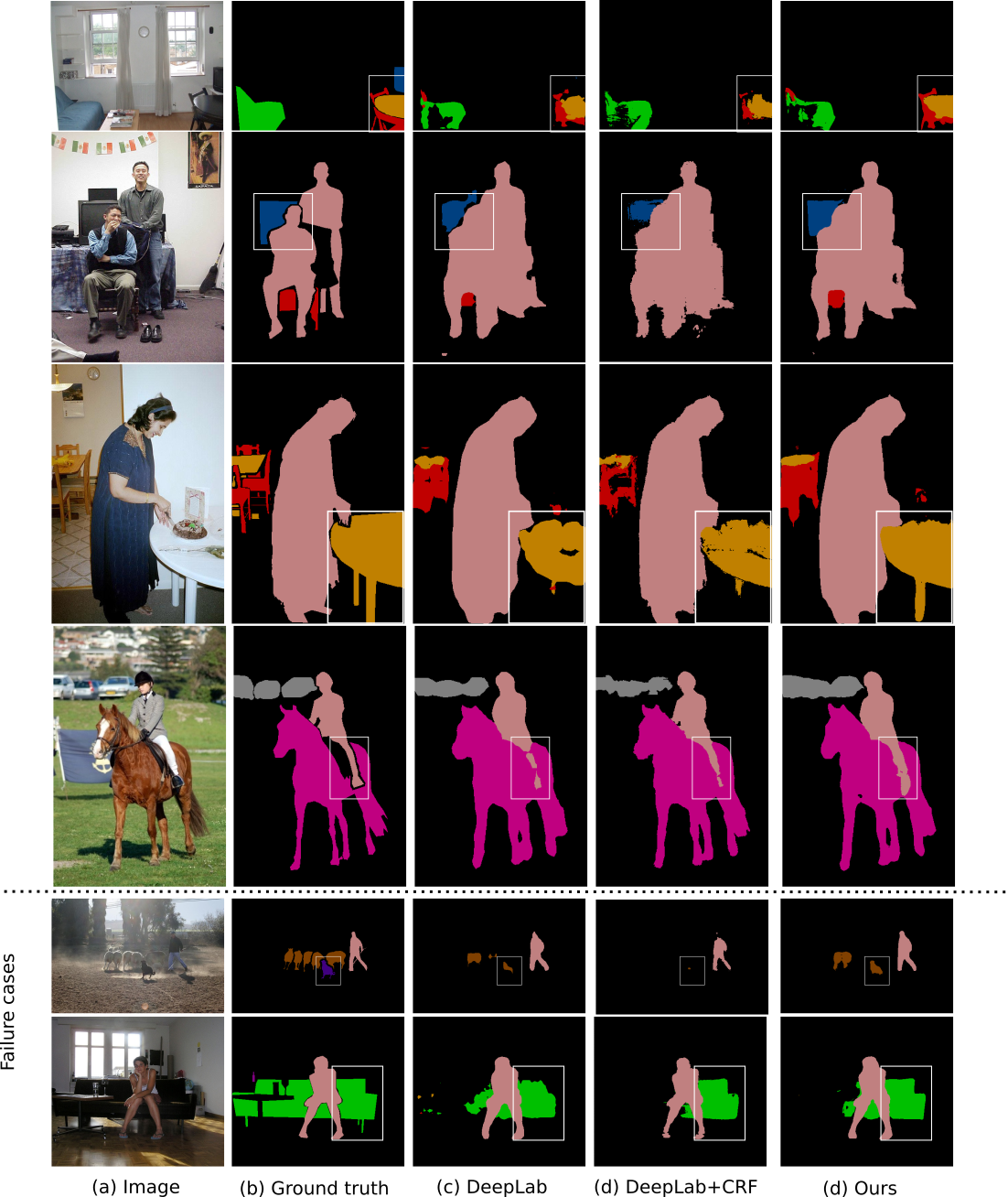}
\end{center}
   \caption{Visualization results on the PASCAL VOC 2012 val set. The first four rows present successful cases while the last two present failure cases. For each row, we present (a) input image, (b) ground truth label, (c) baseline DeepLab-v2 segmentation result, (d) DeepLab-v2+Dense CRF result, (e) segmentation result after applying our full model.}
\label{fig:vis_b_mask}
\end{figure*}

\subsection{Error Analysis}
\label{sec:err_analysis}

In this section, we analyze the improvement our method brings to the object boundaries and discuss its failure cases.

\textbf{Trimap} 
Following previous works \cite{chen2016deeplab, harley2017segmentation}, we quantify the performance near object boundaries. 
For the PASCAL VOC 2012 dataset, we compute the performance on the narrow band ('trimap') near the object boundary. 
Two examples can be found in the left part of Fig.~\ref{fig:trimap}. 
The right part illustrates a plot of mean IoU versus various trimap widths ranging from 1 to 40 pixels. 
It shows that our full model (in blue dotted line) outperforms the baseline and CRF-based method by a certain margin near the object boundaries. 

\textbf{Failure cases} 
Here, we further analyze some failure cases and conclude that
our method can better delineate the boundary but has difficulties in correcting the wrong class labels when a major part of the object is initially wrongly labeled. 
The last two rows in Fig.~\ref{fig:vis_b_mask} illustrate two typical such examples. 
For the first one, our model can better recover the shape of the dog but with the wrong class label.
For the second one, the right side of the sofa is improved, but for the left side the class label is wrongly classified due to occlusion. 

\section{Conclusion}
\label{sec:conclusion}
We have presented two strategies for error correction in dense labeling prediction, and a final model that combines the advantages of these two strategies. 
Our experiments show that our full model improves over state-of-the-art semantic segmentation models for the object semantic segmentation and face parsing tasks.
Compared to other post-processing methods, our approach provides a simpler solution by considering nearby context information for label propagation and at the same time it directly generates new labels for initially wrongly labeled regions. 
In the future, we plan to further reduce the network's size in order to allow for even faster inference times.


\section*{Acknowledgments}
This work was supported by Toyota Motor Europe and the FWO project SFS: Structure From Semantics. 
We would also like to acknowledge the NVIDIA Academic Hardware Grant for providing us with GPUs. 

{\small
\bibliographystyle{ieee}
\bibliography{egbib}
}

\clearpage 
\onecolumn
\setcounter{section}{0}
\renewcommand\thesection{\Alph{section}}
\noindent
{\Large {\textbf{Supplementary materials}}}
\\
\section{Network Architecture}
\label{sec:archi}

Table~\ref{tab:archi} describes the network hyper-parameters for the proposed models in the main paper.
To this point, we want to remind that upon publication the full code including the training details will be made publicly available.

\begin{table*}[h]
\caption{Detailed architecture of the three proposed models. In this table, $E$ denotes our \textit{LabelPropagation} (\textit{Lab.Prop.}) model, $C$ denotes our \textit{LabelReplacement} (\textit{Lab.Repl.)} model, and $M$ denotes our \textit{Fusion} model.}
\begin{center}
\scalebox{0.85}{
\begin{tabular}{|r|l|c c c c| c  |}
\hline
 Name & Kernel  & Str. & Act.& Ch I/O & Input &  Model \\
\hline
 conv1\_1 & 3x3 & 1 & ReLU &(3+\#of classes)/64 & RGB+Seg. & \textit{Lab.Prop.}, \textit{Lab.Repl.} \\
  conv1\_2 & 3x3 & 1 & ReLU &64/64 & conv1\_1 &   \textit{Lab.Prop.}, \textit{Lab.Repl.} \\
  maxpool\_1 & 2x2 & & &64/64 & conv1\_2&   \textit{Lab.Prop.}, \textit{Lab.Repl.}\\
  conv2\_1 & 3x3 & 1 & ReLU & 64/128 & maxpool\_1&   \textit{Lab.Prop.}, \textit{Lab.Repl.}\\
  conv2\_2 & 3x3 & 1 & ReLU & 128/128 & conv2\_1&   \textit{Lab.Prop.}, \textit{Lab.Repl.} \\
 maxpool\_2 & 2x2 &&  & 128/128 & conv2\_2&   \textit{Lab.Prop.}, \textit{Lab.Repl.}\\
 conv3\_1 & 3x3 & 1 & ReLU & 128/256 & maxpool\_2&   \textit{Lab.Prop.}, \textit{Lab.Repl.} \\
 conv3\_2 & 3x3 & 1 & ReLU & 256/256 & conv3\_1&   \textit{Lab.Prop.}, \textit{Lab.Repl.} \\
  maxpool\_3 & 2x2 &&& 256/256 &conv3\_2&   \textit{Lab.Prop.}, \textit{Lab.Repl.}\\
  conv4\_1 & 3x3 & 1 & ReLU & 256/256 & maxpool\_3&   \textit{Lab.Prop.}, \textit{Lab.Repl.}\\
 conv4\_2 & 3x3 & 1 & ReLU & 256/256 &  conv4\_1&   \textit{Lab.Prop.}, \textit{Lab.Repl.}\\
 upsamp1 &  & & & 256/256 & conv4\_2&   \textit{Lab.Prop.}, \textit{Lab.Repl.}\\
 skip1 & & & & 512/512 & upsamp1+conv3\_2&   \textit{Lab.Prop.}, \textit{Lab.Repl.}\\
 E\_conv1\_1 / C\_conv1\_1& 3x3 & 1 & ReLU&512/256 & skip1&   \textit{Lab.Prop.}, \textit{Lab.Repl.} \\
 E\_conv1\_2 / C\_conv1\_2& 3x3 & 1 & ReLU & 256/256& E\_conv1\_1/C\_conv1\_1&   \textit{Lab.Prop.}, \textit{Lab.Repl.} \\
 E\_upsamp2 / C\_upsamp2 & & & & 256/256 & E\_conv1\_2/C\_conv1\_2 &   \textit{Lab.Prop.}, \textit{Lab.Repl.}\\
 E\_skip2/ C\_skip2 & &&& 384/384 & (E\_upsamp2/C\_upsamp2)+conv2\_2 &   \textit{Lab.Prop.}, \textit{Lab.Repl.}\\
 E\_conv2\_1 / C\_conv2\_1 & 3x3 & 1 & ReLU & 384/128 & skip2 &   \textit{Lab.Prop.}, \textit{Lab.Repl.}\\
 E\_conv2\_2 / C\_conv2\_2 & 3x3 & 1 & ReLU & 128/128 & E\_conv2\_1/C\_conv2\_1 &   \textit{Lab.Prop.}, \textit{Lab.Repl.}\\
 E\_upsamp3 / C\_upsamp3 & & & & 128/128 & E\_conv2\_2/C\_conv2\_2  &   \textit{Lab.Prop.}, \textit{Lab.Repl.}\\
 E\_skip3 / C\_skip3 & &&&192/192 & (E\_upsamp3/C\_upsamp3)+conv1\_2 &   \textit{Lab.Prop.}, \textit{Lab.Repl.}\\
 E\_conv3\_1 / C\_conv3\_1 &3x3 &1 &ReLU & 192/64 &skip3 &   \textit{Lab.Prop.}, \textit{Lab.Repl.}\\
 E\_conv3\_2 / C\_conv3\_2 & 3x3 & 1 & ReLU & 64/64 & E\_conv3\_1/C\_conv3\_1 &   \textit{Lab.Prop.}, \textit{Lab.Repl.}\\
 flow & 3x3 & 1&tanh &64/2& E\_conv3\_2 &   \textit{Lab.Prop.}\\
 warp & & & & \#of classes,2/\# of classes & Seg.,flow & \textit{Lab.Prop.}\\ 
 C\_out & 3x3 & 1& &64/\# of classes & C\_conv3\_2&   \textit{Lab.Repl.}\\
 M\_conv1 &3x3 &1 &ReLU & 128/64& E\_conv3\_2+C\_conv3\_2& 
 \textit{Fusion}\\
M\_conv2 &3x3 &1 &ReLU & 64/64&M\_conv1&  \textit{Fusion}\\
M\_conv3 &3x3 &1 &ReLU & 64/256&M\_conv2&  \textit{Fusion}\\
mask & 3x3 & 1 & sigmoid & 256/1 & M\_conv3 & \textit{Fusion}\\ 
combine & & & & (\# of classes)*2,1/\# of classes & C\_out,warp,mask & \textit{Fusion}\\
\hline
\end{tabular}
}
\end{center}
\label{tab:archi}
\end{table*}
\clearpage

\section{Segmentation Results}
\label{sec:vis}
\begin{figure*}[h]
\begin{center}
\includegraphics[width=1\textwidth]{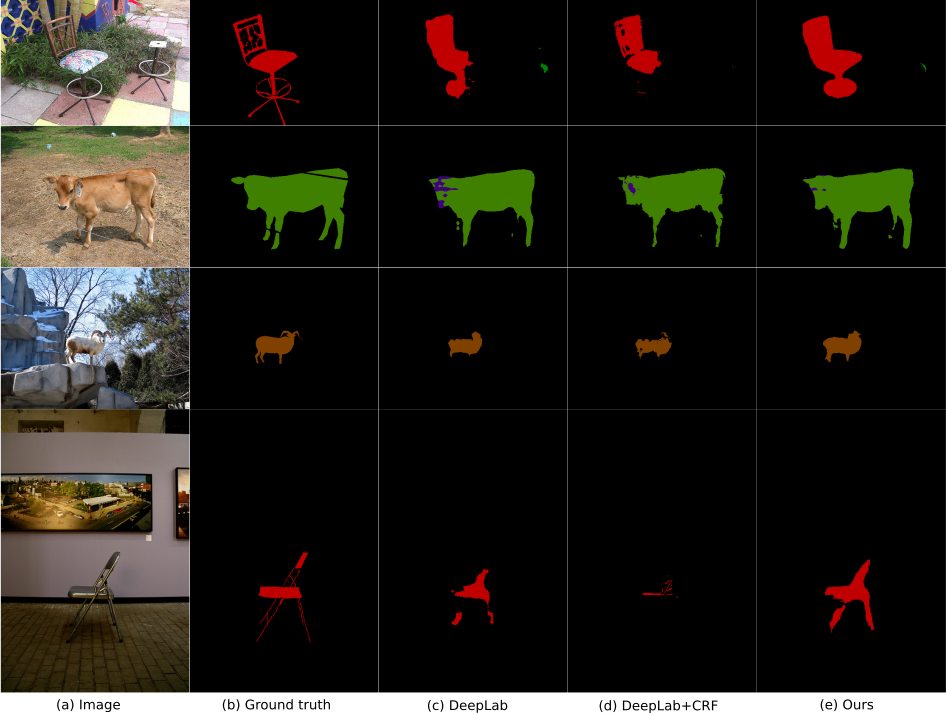}
\end{center}
\caption{Visualization results on the PASCAL VOC 2012 val set. For each row, we present (a) input image, (b) ground truth label, (c) baseline DeepLab-v2 segmentation result, (d) DeepLab-v2+denseCRF result, (e) segmentation result after applying our full model.}
\label{fig:dl}
\end{figure*}

More qualitative results using our models. 
Similar to Fig. 7 in the main paper, in Fig.~\ref{fig:dl} we illustrate more segmentation results using our full model on top of the DeepLab-v2-ResNet initializations.

\begin{figure*}[h]
\begin{center}
\includegraphics[width=1\textwidth]{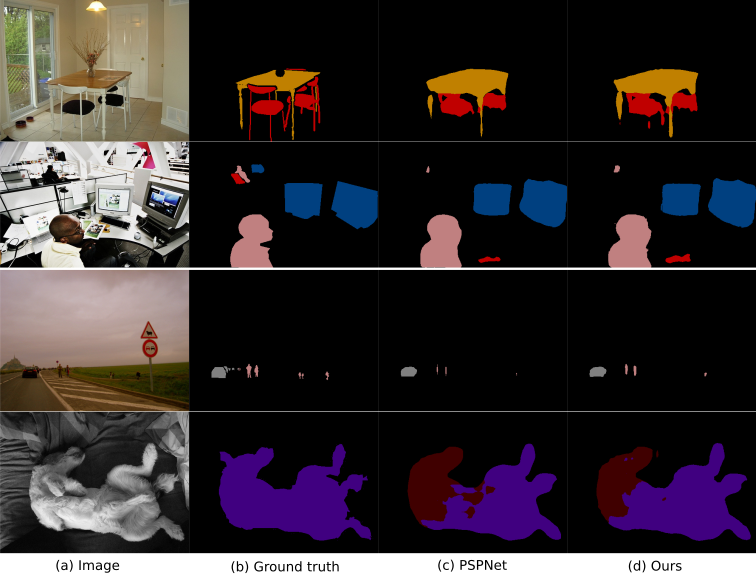}

\end{center}
\caption{Visualization results on the PASCAL VOC 2012 val set. For each row, we present (a) input image, (b) ground truth label, (c) baseline PSPNet segmentation result, (d) segmentation result after applying our full model.}
\label{fig:psp}
\end{figure*}

\clearpage 
Similarly, Fig.~\ref{fig:psp} shows the segmentation results using our full model on top of the PSPNet initializations.

\vspace{1cm}

Visual results of the \textit{LabelPropagation} and \textit{LabelReplacement} networks applied on top of DeepLab-v2-ResNet initializations can be found in Fig.~\ref{fig:flow_cnn} below. Note that, in this case the aforementioned networks are trained and applied independently, whereas for our full model in Fig.~\ref{fig:dl} and Fig.~\ref{fig:psp} they are trained jointly together with the \textit{Fusion} network. 

\begin{figure*}[h]
\begin{center}
\includegraphics[width=0.75\textwidth]{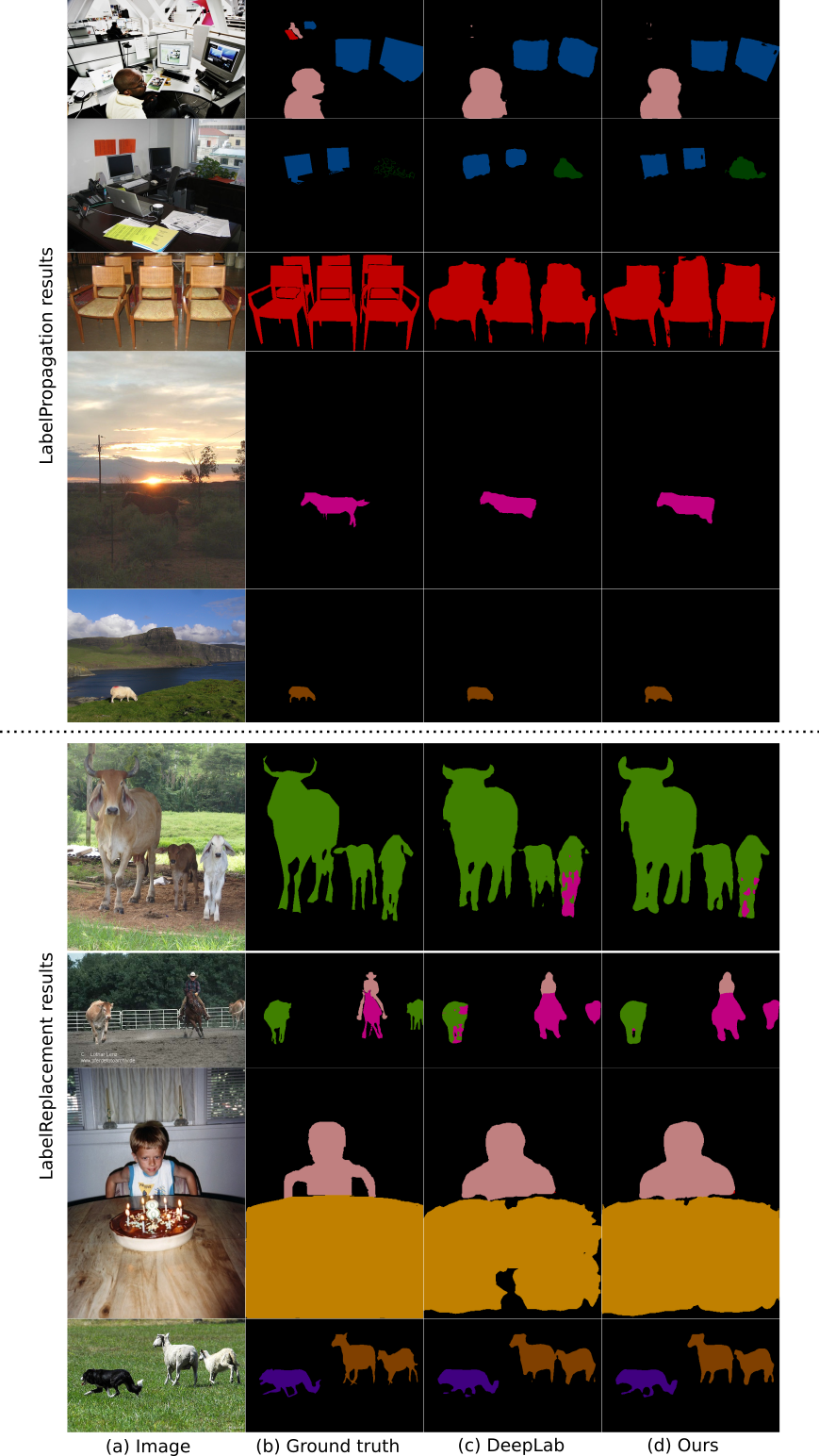}
\end{center}
\caption{Visualization results on the PASCAL VOC 2012 val set. For each row, we present (a) input image, (b) ground truth label, (c) baseline DeepLabv2 segmentation result, (d) segmentation result after applying only our \textit{LabelPropagation} network (top five rows) and only our \textit{LabelReplacement} network (bottom four rows).}
\label{fig:flow_cnn}
\end{figure*}
\end{document}